\documentclass[11pt, a4paper]{article}

\usepackage[utf8]{inputenc}
\usepackage[T1]{fontenc}
\usepackage{geometry}
\usepackage{authblk}      
\usepackage{graphicx}     
\usepackage{booktabs}     
\usepackage{hyperref}     
\usepackage{caption}      
\usepackage{float}        
\usepackage{kotex}        
\usepackage{url}
\usepackage{amsmath}      
\usepackage{multirow}     

\title{\textbf{Implementation of a Skin Lesion Detection System for Managing Children with Atopic Dermatitis Based on Ensemble Learning}}

\author[1]{Soobin Jeon}
\author[2]{Sujong Kim}
\author[1,*]{Dongmahn Seo}

\affil[1]{School of Computer Software, Catholic University of Daegu, Gyeongsan-si, Republic of Korea}
\affil[2]{Department of Software Research, Gyeongbuk Research Institute of Vehicle Embedded Technology, Yeongcheon-si, Republic of Korea}
\affil[*]{Correspondence: sarum@cu.ac.kr; Tel.: +82-10-5106-0388}

\date{} 

\begin{document}

\maketitle

\begin{abstract}
The amendments made to the Data 3 Act and impact of COVID-19 have fostered the growth of digital healthcare market and promoted the use of medical data in artificial intelligence in South Korea. Atopic dermatitis, a chronic inflammatory skin disease, is diagnosed via subjective evaluations without using objective diagnostic methods, thereby increasing the risk of misdiagnosis. It is also similar to psoriasis in appearance, further complicating its accurate diagnosis. Existing studies on skin diseases have used high-quality dermoscopic image datasets, but such high-quality images cannot be obtained in actual clinical settings. Moreover, existing systems must ensure accuracy and fast response times. To this end, an ensemble learning–based skin lesion detection system (ENSEL) was proposed herein. ENSEL enhanced diagnostic accuracy by integrating various deep learning models via an ensemble approach. Its performance was verified by conducting skin lesion detection experiments using images of skin lesions taken by actual users. Its accuracy and response time were measured using randomly sampled skin disease images. Results revealed that ENSEL achieved high recall in most images and $<$1 s processing speed. This study contributes to the objective diagnosis of skin lesions and promotes the advancement of digital healthcare.

\vspace{0.5cm}
\end{abstract}

\section{Introduction}
COVID-19 and the 2022 Russia–Ukraine conflict [1] have primarily driven the rapid expansion of digital healthcare. COVID-19 increased the demand for remote medical consultations and artificial intelligence (AI) diagnostics, whereas the war highlighted the need for sustainable medical services in unstable situations. South Korea introduced strategies to accelerate digital transformation [2] for improving disease management and healthcare services using digital healthcare technologies. Digital healthcare management offers three main benefits. (1) Digital technology enables providing medical services without geographical barriers; (2) data analysis facilitates the early detection of diseases, thereby expanding preventive medicine; and (3) image analysis technology improves disease diagnosis and increases the efficiency of medical consultations. The recent amendment of South Korea’s Data 3 Acts has enabled using medical data for improving healthcare. Thus, researchers and organizations are using such data for disease management research and digital healthcare services.

Implemented in South Korea in 2020, the Data 3 Acts comprises the Personal Information Protection Act, the Act on Promotion of Information and Communications Network Utilization and Information Protection, and the Use and Protection of Credit Information Act [3]. Before these amendments, medical data and biometric information were classified as susceptible personal information and subjected to strict regulations, which limited the use of AI in the country. After these amendments, medical professionals in South Korea have been using AI for skin image analysis and developing various diagnostic solutions.

Atopic dermatitis is a chronic inflammatory skin disease with unclear underlying causes [4]. It is diagnosed by subjectively evaluating lesions based on clinical symptoms, medical history, laboratory findings, and naked-eye observation. This process increases the risk of misdiagnosis, particularly when the observer’s clinical experience is insufficient [5]. Therefore, an ENsemble learning–based Skin Lesion detection system (ENSEL) was proposed herein to enhance the diagnostic accuracy of atopic dermatitis while facilitating an objective and accurate diagnosis for managing atopic diseases in children. Using ENSEL, medical professionals can rely more on data-based objective indicators and less on their subjective judgments, thereby considerably improving the accuracy and reliability of diagnosis.

Atopic dermatitis can be difficult to distinguish from psoriasis because of their similar external characteristics. Therefore, models used for their diagnosis must be trained on diverse skin disease datasets. The majority of existing studies have used dermoscopic image datasets to analyze and diagnose skin lesions. However, obtaining such high-quality images in actual clinical settings is challenging and images taken by users often contain noise, which ultimately reduce the diagnostic accuracy. To address these issues, ENSEL was trained on various skin disease datasets to ensure that it accurately distinguishes atopic dermatitis from other skin conditions.

ENSEL uses the strengths of deep learning models in its ensemble and reduces bias involved when using a single model, thereby improving the generalization performance. The diagnostic accuracy of ENSEL using low-quality images was evaluated by testing it on datasets containing images of different qualities. The diagnostic accuracy, resource usage, and response time of models must be considered for their optimal real-world application across various fields. By optimizing these factors and focusing on balancing their trade-offs, ENSEL can achieve high diagnostic accuracy. To this end, its overall performance was experimentally investigated based on response time and resource usage.

The remainder of this paper is organized as follows. The existing skin disease detection systems and datasets have been reviewed in Section 2, and ENSEL has been introduced in Section 3. Section 4 covers its performance evaluation, and Section 5 concludes the paper and discusses future work.

\section{Related Work}

\subsection{Commercial Software for Skin Disease Detection}
Commercial software such as MoleScope, Aysa, and Skin Vision is used for early diagnosis and management of skin diseases such as skin cancer [6–8]. However, it poses several issues when used in East Asia, particularly South Korea. First, the datasets used to train such software are not publicly disclosed, which raise concerns about its accuracy in diagnosing skin conditions peculiar to Asians. Although this software can effectively diagnose specific skin conditions such as skin cancer, its diagnostic accuracy may be lower for other skin diseases. Additionally, dermoscopic attachments such as those used in MoleScope, which are mounted on smartphones for diagnosing skin conditions, are expensive, making the early diagnosis and detection of skin diseases challenging.

\subsection{Conventional Methodologies for Skin Disease Detection}
Boundary-based and region-based methods have been used for detecting skin diseases from images. However, open datasets such as ISIC and PH2 are primarily and widely used for training AI models [9] because they contain high-quality dermoscopic images with expert labeling and annotations [10–14]. However, diagnostic systems that use models trained on high-quality dermoscopic images may yield low-accuracy results when inferring from real-world images that contain noise and are captured in poor lighting and shooting angles.

\subsection{Real-World Dataset for Skin Disease Detection}
Skin disease datasets specific to population groups reflect the skin characteristics of that population, thereby enabling accurate diagnosis of skin diseases. Researchers in South Korea constructed a medical image dataset containing data on skin conditions in adults and a pediatric skin disease image dataset containing data on skin diseases in children and adolescents, with South Koreans as the target population [15, 16]. Table \ref{tab:table1} shows the class distribution of this medical image dataset containing 106,776 images of skin diseases with their clinical information, including photos of lesions and surrounding skin. This dataset contained 39,493 prospective data entries of overall body photographs, including lesions and the surrounding skin, and 67,283 retrospective data entries processed from existing hospital data by cropping the lesion areas. Table \ref{tab:table2} shows the class distribution for the pediatric skin disease image dataset containing images of 40 types of skin diseases such as atopic dermatitis and normal skin conditions. The dataset contained 8,468 prospective data entries and 49,718 retrospective data entries.

By combining the aforementioned two datasets, skin characteristics across various age groups of population can be determined. These substantial volume of skin image data can be used for model training to accurately detect and diagnose skin diseases.

\begin{table}[ht]
\centering
\caption{Class distribution of medical image datasets for skin disease diagnosis}
\label{tab:table1}
\small
\begin{tabular}{p{0.3\textwidth} p{0.65\textwidth}}
\toprule
\textbf{Category} & \textbf{Conditions} \\
\midrule
Inflammatory Skin Diseases & Atopic dermatitis, Psoriasis, Rosacea, Seborrheic dermatitis, \\
(11 types) & Dermatophytosis, Acne, Folliculitis, Allergic contact dermatitis, Pityriasis rosea, Herpesviral infection \\
\midrule
Skin Tumors & Malignant melanoma, Basal cell carcinoma, Wart, Squamous cell \\
(15 types) & carcinoma, Seborrheic keratosis, Milium, Sebaceous hyperplasia, Hemangioma, Dermatofibroma, Syringoma, Black spot, Soft fibroma, Actinic keratosis, Melanocytic nevus, Epidermal cyst \\
\midrule
Other Common Skin Diseases & Scar, Urticaria, Vitiligo, Idiopathic guttate hypomelanosis, \\
(6 types) & Alopecia areata, Androgenic alopecia \\
\bottomrule
\end{tabular}
\end{table}

\begin{table}[ht]
\centering
\caption{Class distribution of pediatric skin disease image dataset}
\label{tab:table2}
\small
\begin{tabular}{p{0.3\textwidth} p{0.65\textwidth}}
\toprule
\textbf{Category} & \textbf{Conditions} \\
\midrule
Inflammatory Skin Diseases & Atopic dermatitis, Psoriasis, Acne, Insect bite, Nummular eczema, \\
(7 types) & Urticaria, Prurigo \\
\midrule
Congenital Skin Diseases & Melanocytic nevus, Becker nevus, Milk coffee nevus, Salmon patches, \\
(11 types) & Ota like melanosis, Epidermal nevus, Infantile hemangioma, Nevus sebaceous, Congenital melanocytic nevus, Capillary malformation, Mongolian spot \\
\midrule
Infectious Skin Diseases & Varicella, Herpes simplex, Molluscum contagiosum, Dermatophytosis, \\
(7 types) & Wart, Impetigo, Pityriasis versicolor \\
\midrule
Other Skin Diseases & Nail dystrophy, Pyogenic granuloma, Ingrown nails, Onychomycosis, \\
(15 types) & Vitiligo, Juvenile xanthogranuloma, Alopecia areata, Melanonychia, Keloid, Mastocytoma, Pityriasis lichenoides chronica, Epidermal cyst, Nevus depigmentosus, Scar, Lichen striatus \\
\bottomrule
\end{tabular}
\end{table}

\section{Proposed System Architecture}

\begin{figure}[htbp]
  \centering
  \includegraphics[width=0.8\textwidth]{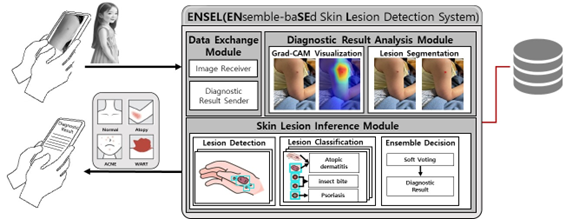}
  \caption{Architecture of the ensemble learning–based skin lesion detection system}
  \label{fig:fig1}
\end{figure}

Figure \ref{fig:fig1} shows the proposed ENSEL architecture. It comprises a data exchange module, diagnostic result analysis module, and skin lesion inference module. The data exchange module contains an image receiver that receives images from users and a diagnostic result sender that conveys the lesion diagnosis results to users. The diagnostic result analysis module provides heatmap visualizations for system administrators to identify areas focused on by the classification model using gradient-weighted class activation mapping (Grad-CAM). Lesion segmentation visualizes the lesion areas analyzed by the skin lesion inference module. The skin lesion inference module diagnoses skin diseases from real-world images sent by users via lesion detection, classification, and ensemble decision. For lesion detection, the module performs localization to identify the size and position of lesions within an image. In lesion classification, the module classifies the overall image and the localized lesion areas using various trained classification models. The results of each lesion classification are then integrated to make a final diagnosis during ensemble decision. Soft voting is used to predict the probability for each lesion using classification models, and a final classification outcome is achieved.

\begin{figure}[htbp]
  \centering
  \includegraphics[width=0.8\textwidth]{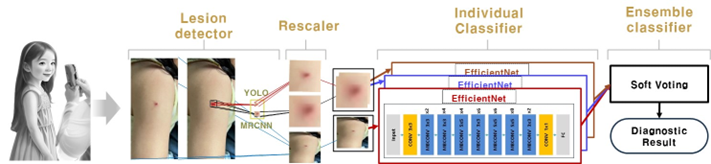}
  \caption{Diagnostic flowchart of ENSEL for detecting skin diseases}
  \label{fig:fig2}
\end{figure}

Figure \ref{fig:fig2} shows the diagnostic flowchart of ENSEL for detecting skin diseases. It comprises the lesion detector, rescaler, individual classifier, and ensemble classifier. The lesion detector uses multiple detection models to pinpoint the locations of skin lesions. The rescaler modifies the dimensions of lesion images identified by the lesion detector to ensure they are optimal for the subsequent phase, i.e., the individual classifier. The individual classifier uses various classification models to analyze and evaluate each lesion independently. The ensemble classifier consolidates the analysis results from the individual classifiers using the soft voting method to aggregate the results of each classification model and obtain final diagnosis.

\subsection{Data Exchange Module}
The data exchange module processes image data sent by users and transmits the ENSEL-analyzed diagnostic results to users. This module comprises an image receiver and diagnostic result sender. The image receiver receives skin lesion images uploaded by users via desktop or mobile devices. It then converts the received images into a format suitable for processing and analysis by the system via basic preprocessing tasks such as format conversion. The diagnostic result sender transmits the analyzed results from ENSEL to users. The diagnostic results are provided in text and image formats, where the text includes the name and probability values of the diagnosed skin diseases. The images visually indicate the location of lesions found by ENSEL using alpha blending.

\subsection{Diagnostic Result Analysis Module}

\begin{figure}[htbp]
  \centering
  \includegraphics[width=0.8\textwidth]{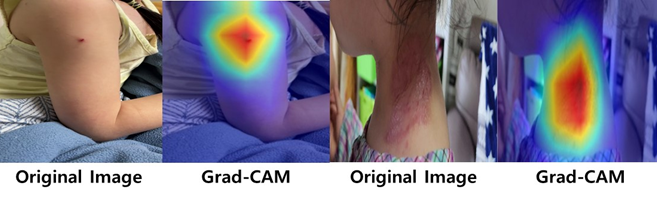}
  \caption{Example of Grad-CAM visualization applied to skin disease images}
  \label{fig:fig3}
\end{figure}

The diagnostic result analysis module analyzes the diagnostic process of ENSEL and visualizes the skin lesions. This module comprises Grad-CAM visualization and lesion segmentation. Grad-CAM visualization provides a heatmap of the areas that ENSEL focused on during skin disease diagnosis. This visualization allows administrators using ENSEL to review and analyze the areas emphasized during diagnosis, enabling further adjustments or tuning to improve the model performance. Figure \ref{fig:fig3} shows an example of a skin disease image obtained after applying Grad-CAM visualization. Lesion segmentation visualizes the location of lesions identified by ENSEL, helping to determine the exact size, shape, and location of the skin lesions. This visualization aids users to visually understand the diagnostic results obtained from ENSEL, thereby increasing the trust in diagnosis.

\subsection{Skin Lesion Inference Module}

\begin{figure}[htbp]
  \centering
  \includegraphics[width=0.8\textwidth]{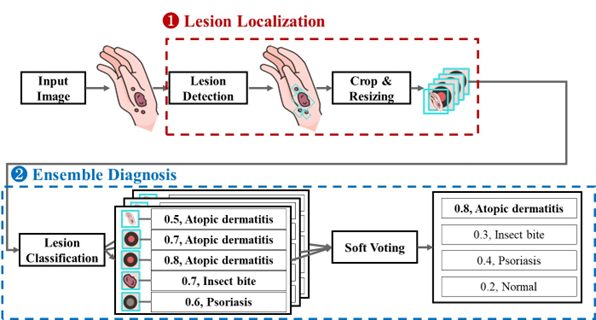}
  \caption{Diagnostic flowchart of the skin lesion inference module}
  \label{fig:fig4}
\end{figure}

Figure \ref{fig:fig4} shows the diagnostic flow of the skin lesion inference module that includes the lesion localization and ensemble diagnosis stages. In the lesion localization stage, the system identifies the location of the actual lesion areas in an image and crops them. It resizes them to prepare the data for further processing. In the ensemble diagnosis stage, various classification models yield separate predictions, which are then combined to reach the final diagnosis. Each model calculates probability values for various skin diseases such as atopic dermatitis, insect bites, and psoriasis. The system uses soft voting method to aggregate these probabilities and quantifies the likelihood of each disease, thereby producing the final diagnosis results. This approach improves the accuracy of AI-based skin disease diagnosis and addresses the limitations of relying on the judgment of a single model.

\subsubsection{Selection and Application of Deep Learning Models}
Herein, we use mask region-based convolutional neural network (Mask R-CNN) [17] and YOLOv8 [18] for lesion localization and EfficientNet [19] for ensemble diagnosis. Region-based R-CNN, which offers high precision despite its slow speed, and regression-based YOLO, which is faster but slightly less accurate, are commonly used deep learning algorithms [20]. Mask R-CNN, an extension of faster R-CNN, can detect objects and perform pixel-level segmentation. YOLOv8, one of the most recent models in the YOLO series, shows higher processing speed and accuracy than its predecessors.

Vision transformers (ViTs) have recently shown high performance in computer vision tasks and are emerging as alternatives to CNNs [21]. However, CNN-based EfficientNet can be trained on small datasets contrary to ViTs. A comparison of the performance of various models, including ViT and EfficientNet, trained on CIFAR-10, CIFAR-100, SVHN, and T-ImageNet datasets, revealed that EfficientNet exhibited higher overall accuracy and image throughput per second than ViT [22–25]. As the real-world dataset for skin disease detection is small, EfficientNet was used for lesion classification herein.

\subsubsection{Deep Learning Model Training}
ENSEL was trained on real-world datasets: medical image dataset for skin disease diagnosis [15] and pediatric skin disease image dataset [16]. These datasets were divided into prospective and retrospective data. Mask R-CNN and YOLOv8 used prospective data, whereas EfficientNet used retrospective data. Mask R-CNN and YOLOv8 focused on identifying the location and shape of lesions for accurate localization. Models can develop precise disease localization capabilities by clearly learning the distinction between lesions and normal skin within the datasets. EfficientNet learns data for 40 types of skin diseases, including atopic dermatitis, and healthy skin using retrospective data from both datasets.

The prospective data in the medical image dataset for skin disease diagnosis comprised 39,493 images, which were divided into 27,645 images for training, 7,898 images for validation, and 3,950 images for testing. The retrospective data, comprising 67,283 images, was divided into 47,098 images for training, 13,456 images for validation, and 6,729 images for testing. The prospective data in the pediatric skin disease image dataset contained 8,468 images, which were divided into 6,774 images for training, 847 images for validation, and 847 images for testing. The retrospective data containing 49,718 images were divided into 39,774 images for training, 4,972 images for validation, and 4,972 images for testing.

The deep learning model was trained in three phases. In August 2022, prospective and retrospective data from the medical image dataset for skin disease diagnosis were used to train Mask R-CNN and EfficientNet, respectively, in the first phase. In February 2023, prospective data and retrospective data from the pediatric skin disease image dataset were used to train YOLOv8 and EfficientNet, respectively. Finally, in July 2023, both datasets were used to improve the classification performance of EfficientNet via model training and performance evaluation.

\subsubsection{Evaluation of Deep Learning Model Performance}

\begin{figure}[htbp]
  \centering
  \includegraphics[width=0.6\textwidth]{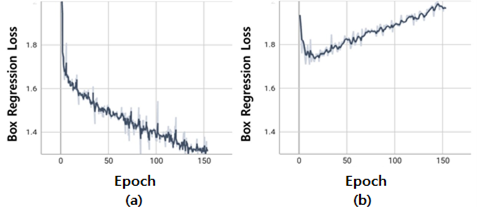}
  \caption{Mask R-CNN box regression loss graph: (a) train box regression loss and (b) validation regression loss.}
  \label{fig:fig5}
\end{figure}

Figure \ref{fig:fig5} shows the changes in the box regression loss during the training and validation of Mask R-CNN. The box regression loss is a loss function used to calculate the difference between the predicted and actual bounding boxes. As shown in Figure 5(a), the loss decreases over time, indicating that the model is learning the characteristics of skin diseases. However, the validation graph in Figure 5(b) shows irregular fluctuations, suggesting differences between the training and validation datasets. Therefore, the model was selected at an appropriate epoch to prevent overfitting.

\begin{figure}[htbp]
  \centering
  \includegraphics[width=0.6\textwidth]{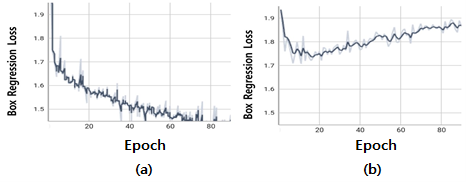}
  \caption{YOLOv8n box regression loss graph: (a) train box regression loss and (b) validation regression loss.}
  \label{fig:fig6}
\end{figure}

Figure \ref{fig:fig6} shows the changes in the box regression loss during the training and validation of YOLOv8n. As shown in Figure 6(a), the loss continuously decreases as the epochs progress. However, Figure 6(b) shows that the loss increases as the epochs progress, indicating model overfitting. An early stopping technique was applied to select a model that avoids overfitting to ensure the generalization ability of the model.

The performance of EfficientNet trained on various datasets was evaluated using the following four metrics. Precision, which gauges the model’s ability to identify healthy individuals correctly, serves as a measure of false positives. Recall, the inverse of precision, assesses the accuracy of the model in identifying individuals with diseases, serving as a measure of false negatives. The F1-score combines precision and recall and is a single comprehensive metric. Accuracy, the proportion of correct predictions out of all predictions, was also considered.

\begin{table}[ht]
\centering
\caption{Class distribution of medical image dataset for skin disease diagnosis}
\label{tab:table3}
\resizebox{\textwidth}{!}{
\begin{tabular}{lllcclll}
\toprule
ID & Training Phase & Model Version & Class Count & Precision & Recall & F1-score & Accuracy \\
\midrule
M1 & 1st & EfficientNet B7 & 6 & 0.34 & 0.54 & 0.42 & 0.32 \\
M2 & 2nd & EfficientNet B0 & 41 & 0.94 & 0.72 & 0.81 & 0.85 \\
M3 & 2nd & EfficientNet B0 & 41 & 0.92 & 0.5 & 0.64 & 0.75 \\
M4 & 2nd & EfficientNet B0 & 6 & 0.90 & 0.54 & 0.67 & 0.76 \\
M5 & 2nd & EfficientNet B7 & 6 & 0.82 & 0.64 & 0.71 & 0.77 \\
M6 & 3rd & EfficientNet B0 & 4 & 0.81 & 0.72 & 0.76 & 0.80 \\
M7 & 3rd & EfficientNet B0 & 4 & 0.73 & 0.78 & 0.75 & 0.77 \\
M8 & 3rd & EfficientNet B0 & 4 & 0.85 & 0.82 & 0.83 & 0.85 \\
M9 & 3rd & EfficientNet B0 & 4 & 0.51 & 0.44 & 0.47 & 0.55 \\
\bottomrule
\end{tabular}
}
\end{table}

Table \ref{tab:table3} compares the performance of EfficientNet models trained on the skin disease dataset in three phases according to the metrics. Model M1 shows relatively low performance than the models trained in subsequent phases. M1 exhibited low performance, possibly because it was trained on limited training data. Model M2, trained in the second phase using retrospective data, showed high precision and recall, indicating that pediatric skin disease image data considerably improved its disease classification ability. High precision is crucial for accurately classifying diseases, which is a critical factor in medical diagnosis. Model M8, which utilized comprehensive data from both datasets, demonstrated improved classification performance, highlighting the importance of data diversity and quantity in model performance.

\begin{figure}[htbp]
  \centering
  \includegraphics[width=0.7\textwidth]{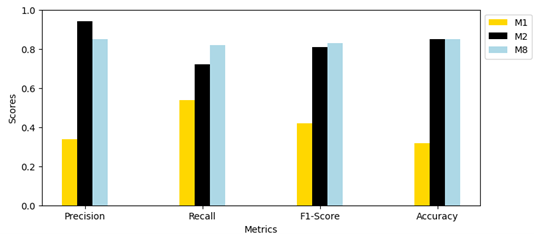}
  \caption{Comparison of the best-performing EfficientNet models by phase.}
  \label{fig:fig7}
\end{figure}

Figure \ref{fig:fig7} compares the best-performing EfficientNet models from each phase according to the metrics. Model M2 excels in precision, which is crucial for reducing false positives. Model M8 excels in recall, indicating its superior ability to detect actual disease samples accurately.

\subsubsection{Application of Ensemble Methods}
Ensemble methods combine multiple models to overcome the limitations or errors of a single model [26]. Soft voting is an ensemble method that averages the predicted probabilities of multiple models and selects the class with the highest average probability [27]. It improves the overall prediction performance of the model by reducing the impact of errors or biases from individual models. Herein, soft voting was applied based on the prediction results of each model. It combined the probability values predicted by each model and selected the class with the highest probability as the final diagnostic result. This approach achieved more reliable predictions by integrating the results of multiple models rather than relying on the prediction of a single model.

Soft voting is a simple aggregation method based on the majority rule. However, it combines information from different models and reflects the confidence levels of each model’s predictions, making it particularly advantageous. In complex medical problems such as skin disease diagnosis, utilizing the multidimensional information provided by various models is crucial. Soft voting considers the predicted probabilities from each model, enabling more accurate and detailed diagnostic decisions. It thus reduces error rates and enhances diagnostic accuracy compared with a single model.

Skin lesion detection experiments were conducted using images of various skin lesions in actual users to validate the improved predictive performance of ENSEL.

\section{System Implementation and Experimental Results}

\subsection{System Environment}
Table \ref{tab:table4} shows the development environment for ENSEL that uses Python as the primary programming language. GPU acceleration leverages NVIDIA’s CUDA and cuDNN for data processing and model training. The API server and a simple user interface were built using Django, and MariaDB handles database management. Additional image processing was performed using OpenCV and NumPy. The overall system was developed in an Ubuntu 18.04 environment.

\begin{table}[ht]
\centering
\caption{ENSEL development environment}
\label{tab:table4}
\begin{tabular}{ll}
\toprule
\textbf{Category} & \textbf{Development Environment/Version} \\
\midrule
language & Python 3.7 \\
deep learning tools & CUDA 10.1, cuDNN 7.6.4 \\
deep learning framework & Pytorch 1.10.1 \\
web framework & Django 3.2.22 \\
operating system & Ubuntu 18.04 \\
database & MariaDB 10.3 \\
others & OpenCV 4.7.0.68, NumPy 1.21.6 \\
\bottomrule
\end{tabular}
\end{table}

\subsection{System Implementation}
ENSEL uses a web-based interface, allowing users to access and diagnose skin lesions easily. It accepts original images and uses deep learning models to provide lesion results via an API server developed with Python and Django. This server integrates and optimizes deep learning models using PyTorch. MariaDB manages user data and diagnostic results. The system architecture follows a client–server model, with the client implemented as a web application or a mobile app (Android or iOS). A simple web client was also developed to enable users to upload skin lesion images and view diagnostic results. The API server receives the uploaded images, preprocesses them using OpenCV and NumPy, and analyzes them with PyTorch-based deep learning models. ENSEL enhances the diagnostic speed and accuracy by optimizing model computations with CUDA and cuDNN.

\begin{figure}[htbp]
  \centering
  \includegraphics[width=0.8\textwidth]{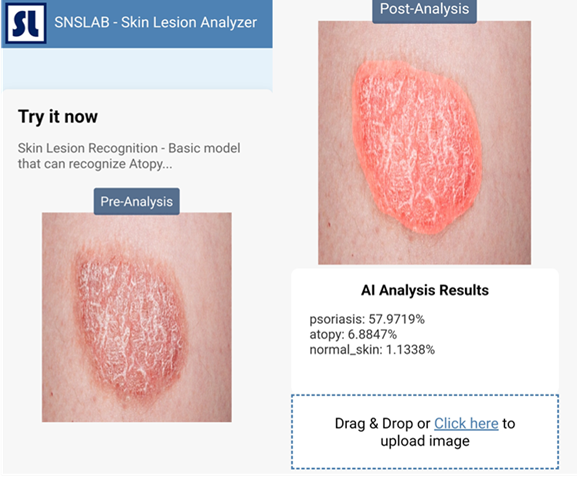}
  \caption{ENSEL client user interface}
  \label{fig:fig8}
\end{figure}

Figure \ref{fig:fig8} shows the ENSEL client user interface, which shows the uploaded image, AI-based skin disease analysis results, and AI-predicted skin lesion area visualization. Users can upload skin lesion images via drag-and-drop or click, initiating ENSEL’s analysis. The analysis results appear in the AI analysis results section, providing users with specific probabilities and detailed metrics for each disease. The AI-predicted skin lesion area visualization highlights the detected lesion locations on the image, allowing users to verify the system’s predictions.

\subsection{Ensemble Deep Learning–Based Skin Lesion}
ENSEL was developed for managing atopic dermatitis in children, as well as improve diagnostic accuracy and achieve a more objective diagnosis. ENSEL aids in objectively diagnosing atopic dermatitis and other skin conditions with complex patterns and varied characteristics. Its ability to accurately detect and classify various skin lesions and its practicality were experimentally evaluated. As ENSEL operates on a cloud infrastructure, the costs and response times of running high-performance deep learning models were evaluated to optimize resource usage.

\subsubsection{Experimental Environment and Method}
Table \ref{tab:table5} shows the experimental environment. Experiments were conducted using 110 randomly sampled images of various skin conditions, including atopic dermatitis, from our dataset of 983 images. Table \ref{tab:table6} shows the resolution of each skin condition image. 110 images were uploaded to ENSEL, and its highly accurate diagnostic results were compared with the actual diagnosis to evaluate its performance. Experiments on response time were conducted to measure the total processing time from uploading each image to delivering the diagnostic result to the user.

\begin{table}[ht]
\centering
\caption{Experimental environment}
\label{tab:table5}
\begin{tabular}{ll}
\toprule
\textbf{H/W} & \textbf{Specification} \\
\midrule
CPU & Intel(R) Xeon(R) Silver 4114 CPU @ 2.20GHz x 2 \\
GPU & Nvidia Tesla V100-PCIE 32GB x 2 \\
RAM & 64GByte \\
\bottomrule
\end{tabular}
\end{table}

\begin{table}[ht]
\centering
\caption{Experimental image resolutions and quantities}
\label{tab:table6}
\begin{tabular}{ll}
\toprule
\textbf{Category} & \textbf{Quantity} \\
\midrule
Below & 7 \\
1,080 $\times$ 1,920 & 40 \\
3,000 $\times$ 4,000 & 12 \\
3,024 $\times$ 4,032 & 4 \\
3,840 $\times$ 2,160 & 46 \\
\bottomrule
\end{tabular}
\end{table}

\subsubsection{Ensemble Deep Learning–Based Skin Lesion Detection Experiment and Results}
The ensemble deep learning–based skin lesion detection experiment was conducted to evaluate the performance of the best single model and ENSEL using quantitative metrics such as precision, recall, F1-score, and accuracy. 110 images were uploaded to ENSEL, and its diagnostic results were compared with the actual diagnostic results to determine its performance. Table \ref{tab:table7} compares the top-performing single models in each phase and the first and second-phase ENSELs that use these models. The first-phase ENSEL includes the best-performing single models that were trained up to the second phase, whereas the second-phase ENSEL includes the highest-performing models from all training phases.

\begin{table}[ht]
\centering
\caption{Performance comparison between single model and ENSEL}
\label{tab:table7}
\resizebox{\textwidth}{!}{
\begin{tabular}{ccccccc}
\toprule
\textbf{Model} & \textbf{Training} & \textbf{Model} & \multirow{2}{*}{\textbf{Precision}} & \multirow{2}{*}{\textbf{Recall}} & \multirow{2}{*}{\textbf{F1-score}} & \multirow{2}{*}{\textbf{Accuracy}} \\
\textbf{Type} & \textbf{Phase} & \textbf{Combination} & & & & \\
\midrule
\multirow{4}{*}{\shortstack{single\\model}} & 1st & M1 & 0.34 & 0.54 & 0.42 & 0.32 \\ \cmidrule{2-7}
 & \multirow{2}{*}{2nd} & M2 & 0.94 & 0.72 & 0.81 & 0.85 \\
 & & M4 & 0.90 & 0.54 & 0.67 & 0.76 \\ \cmidrule{2-7}
 & 3rd & M8 & 0.85 & 0.82 & 0.83 & 0.85 \\
\midrule
\shortstack{first\\ENSEL} & \shortstack{cross\\phase} & M2 + M4 & 0.93 & 0.82 & 0.87 & 0.89 \\
\midrule
\shortstack{second\\ENSEL} & \shortstack{cross\\phase} & M2 + M8 & 0.85 & 0.98 & 0.91 & 0.91 \\
\bottomrule
\end{tabular}
}
\end{table}

\begin{figure}[htbp]
  \centering
  \includegraphics[width=0.7\textwidth]{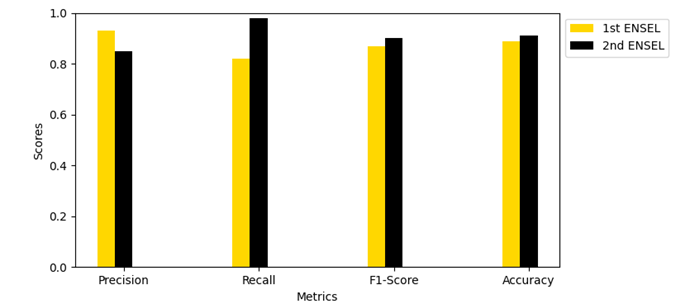}
  \caption{Performance comparison of first- and second-phase ENSEL.}
  \label{fig:fig9}
\end{figure}

Figure \ref{fig:fig9} compares the performance of the first- and second-phase ENSELs. The former shows higher precision, whereas the latter performs better in all other metrics. The second-phase ENSEL excels in recall, a key metric that indicates its exceptional ability to identify individuals with actual skin conditions without false negatives. This high recall is particularly crucial in medical fields, where missing symptoms are unacceptable. It also confirms the effectiveness of ENSEL in identifying skin conditions.

\begin{figure}[htbp]
  \centering
  \includegraphics[width=0.7\textwidth]{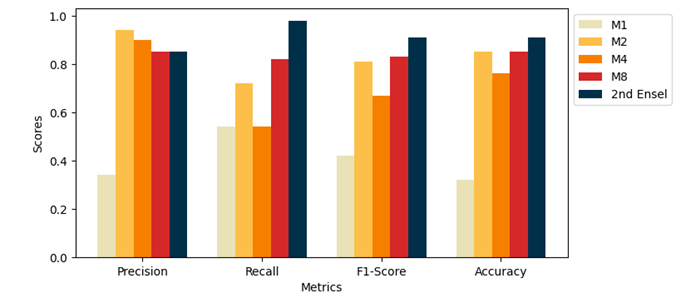}
  \caption{Performance comparison of single models and 2nd ENSEL.}
  \label{fig:fig10}
\end{figure}

Figure \ref{fig:fig10} compares the performance of single models and the second-phase ENSEL. The second-phase ENSEL consistently outperforms single models, particularly in recall and F1-score, indicating that it detects actual lesions more effectively; it thus considerably enhances the accuracy and reliability of medical diagnosis. These findings confirm that ENSEL achieves high performance by integrating the predictions of various models, overcoming the limitations of single models.

GradCAM plays a crucial role in improving the interpretability of deep learning model predictions. By visually showing the parts of an input image that influence the predictions, GradCAM facilitates better understanding of the model behavior [28]. Herein, GradCAM was used to visualize the areas considered important by each single model during prediction and examine how these areas complement each other. Specific models trained on skin conditions may focus on the edges of lesions, whereas others emphasize the center. Poorly trained models may also highlight nonrelevant areas. This process ensures the reliability of models and their potential to improve medical image analysis.

\begin{figure}[htbp]
  \centering
  \includegraphics[width=0.9\textwidth]{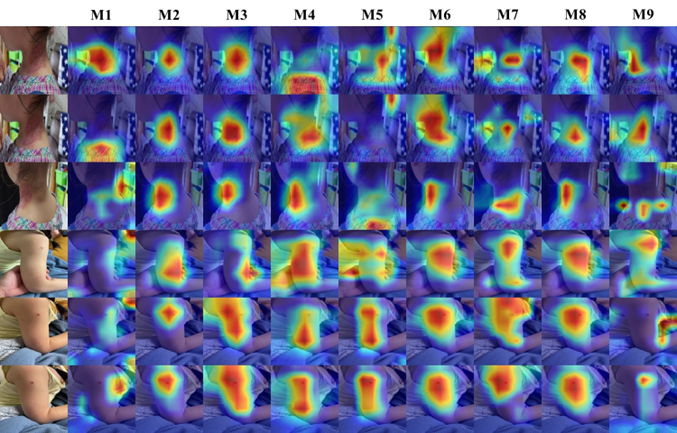}
  \caption{Comparison of activation areas of skin lesions in each single model using GradCAM.}
  \label{fig:fig11}
\end{figure}

Figure \ref{fig:fig11} shows the activation levels in different areas for multiple single models during skin lesion detection, as visualized by GradCAM. The intensity of these colors denotes the importance each model places on certain parts of the image for lesion detection. Models with M2 and M8 show activations spanning the center and edges of lesions, indicating that they comprehensively use the overall area of the lesion for diagnosis. GradCAM demonstrates that each model in the ensemble recognizes different lesion characteristics, and their combination in the ensemble model compensates for shortcomings of the single. Herein, M2 and M8 were combined to enhance the diagnostic capabilities and address their respective limitations.

\subsubsection{System Response Time Experiment and Results}
The system response time experiment was conducted to measure response times accurately, such that the costs associated with AWS usage can be minimized by modifying the system structure. The total processing time from when an image is uploaded until the diagnostic result reaches the user was measured. A total of 220 measurements were collected by uploading 110 images twice.

\begin{figure}[htbp]
  \centering
  \includegraphics[width=0.8\textwidth]{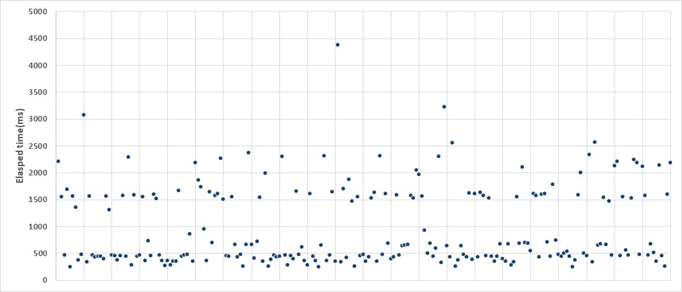}
  \caption{Distribution of system response times.}
  \label{fig:fig12}
\end{figure}

Figure \ref{fig:fig12} shows the distribution of system response times, revealing that 140 uploads were processed in $<$1 s and the remaining 80 uploads were processed in $>$1 s. Although most uploads were processed relatively quickly, many required longer processing times. Table \ref{tab:table7} shows a considerable variation between the minimum and maximum response times. The average system response time is 907 ms, indicating its relatively fast processing. However, the minimum and maximum response times are 264 and 4,386 ms, respectively, indicating that the system sometimes required considerably longer processing times. This significant time discrepancy indicates a need for system improvements to maintain consistent performance.

\begin{figure}[htbp]
  \centering
  \includegraphics[width=0.9\textwidth]{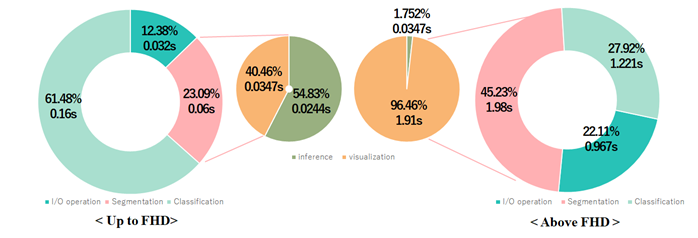}
  \caption{Analysis of image segmentation times in low- and high-resolution images.}
  \label{fig:fig13}
\end{figure}

Figure \ref{fig:fig13} shows the image segmentation processing times of ENSEL for handling the lowest- and highest-resolution images. For low-resolution images, the model inference time is higher than the visualization time during image segmentation. However, for high-resolution images, the visualization time is considerably higher than the inference time. These findings show that visualization consumes substantial system resources when processing high-resolution images. Therefore, visualization must be improved for system optimization and performance enhancement.

\begin{figure}[htbp]
  \centering
  \includegraphics[width=0.9\textwidth]{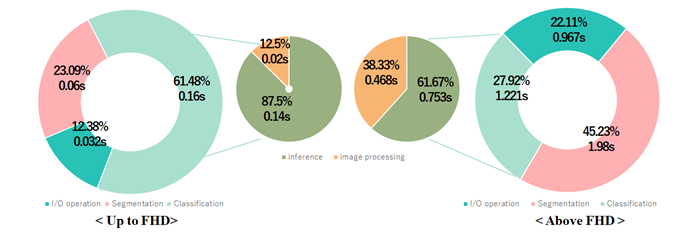}
  \caption{Analysis of image classification times in low- and high-resolution images}
  \label{fig:fig14}
\end{figure}

Figure \ref{fig:fig14} shows the image classification processing times of ENSEL for lowest- and highest-resolution images. For low-resolution images, model inference consumed 87.5\% of the system processing time and the remaining 12.5\% was spent on image processing tasks. In contrast, inference constituted 61.67\% of the time for high-resolution images, whereas image processing tasks accounted for 38.33\%. These results indicate that high image resolution considerably impacts processing time.

Additionally, most images were processed in $<$1 s. However, significant delays were observed when handling high-resolution images. This issue became particularly pronounced when alpha blending was applied to highlight the expected lesion areas for the user, likely due to increased number of pixels for processing. High-resolution images required significant time for processing because of their resolution and increased system complexity caused by specific image processing techniques such as alpha blending. Thus, high image resolution increases processing time, indicating some processing steps must be optimized to improve the overall system performance.

\section{Conclusions}
Atopic dermatitis is a chronic inflammatory skin disease with unclear fundamental causes. At present, objective methods have not been developed for diagnosing this skin conditions, and clinicians rely on clinical symptoms, patient history, laboratory findings, and subjective evaluations via visual inspection. The likelihood of misdiagnosis increases with insufficient clinical experience of the observer. To address these issues, ENSEL was proposed herein for diagnosing atopic dermatitis and assisting observers in making objective diagnosis. The performance of ENSEL was verified by experimentally determining its lesion detection accuracy and response time. 110 images of various skin diseases that were randomly sampled were used for the experiments. ENSEL exhibited superior performance to single models trained on skin disease image datasets. Its recall was 0.98, indicating its capability to accurately detect actual skin diseases without missing most cases. Moreover, ENSEL processed 68\% of the images in $<$1 s, indicative of its high processing speed. However, the remaining 32\% exceeded 1 s because relatively long processing times were required for visualizing high-resolution images. This indicated that its performance varied for different image resolutions. In future, ENSEL will be commercialized on actual cloud-based infrastructure. Moreover, experiments will be conducted to reduce the response times and enhance the stability of ENSEL.

\section*{Supplementary Materials}

\section*{Institutional Review Board Statement}

\section*{Informed Consent Statement}

\section*{Data Availability Statement}

\section*{Acknowledgments}
The National Research Foundation of Korea (NRF) grant supported this work.

\section*{Conflicts of Interest}
The authors declare no conflicts of interest.


\end{document}